\documentclass[authoryear,preprint,review,12pt]{elsarticle}

\usepackage{amsmath,amssymb,amsfonts}
\usepackage{algorithm}
\usepackage{algpseudocode}
\usepackage{setspace}
\usepackage{graphicx}
\usepackage{textcomp}
\usepackage{array}
\usepackage{stfloats}
\usepackage{url}
\usepackage{verbatim}
\usepackage{multirow}

\usepackage[pagebackref,breaklinks,colorlinks,citecolor=cyan,linkcolor=cyan,bookmarks=false]{hyperref}

\newcommand{\be}{\begin{eqnarray}}
\newcommand{\ee}{\end{eqnarray}}
\newcommand{\bee}{\begin{eqnarray*}}
\newcommand{\eee}{\end{eqnarray*}}

\newcommand{\matrixb}{\left[ \begin{array}}
\newcommand{\matrixe}{\end{array} \right]}

\newcommand{\app}{\raise.17ex\hbox{$\scriptstyle\sim$}}

\usepackage{xspace}
\makeatletter
\def\onedot{\ifx\@let@token.\else.\null\fi\xspace}
\makeatother

\makeatletter
\newcommand*\bigcdot{\mathpalette\bigcdot@{.5}}
\newcommand*\bigcdot@[2]{\mathbin{\vcenter{\hbox{\scalebox{#2}{$\m@th#1\bullet$}}}}}
\makeatother

\def\eg{\textit{e.g}\onedot} 
\def\ie{\textit{i.e}\onedot}

\newcommand{\Cref}[1]{Chap.~\ref{#1}}

\usepackage{algorithm}
\usepackage{algpseudocode}

\usepackage[normalem]{ulem}
\usepackage{booktabs}
\usepackage{natbib}
\renewcommand{\cite}{\citep}

\def\BibTeX{{\rm B\kern-.05em{\sc i\kern-.025em b}\kern-.08em
    T\kern-.1667em\lower.7ex\hbox{E}\kern-.125emX}}

\journal{Medical Image Analysis}

\begin{document}

\begin{frontmatter}

\title{Spatial-Temporal Pre-Training for Embryo Viability Prediction Using Time-Lapse Videos} 

Zhiyi Shi, Junsik Kim, Helen Y. Yang, Yonghyun Song, Hyun-Jic Oh,\\Dalit Ben-Yosef, Daniel Needleman, and Hanspeter Pfister

\author[label1]{Zhiyi Shi\textsuperscript{*}} 
\author[label1]{Junsik Kim\textsuperscript{*}} 
\author[label2]{Helen Y. Yang}
\author[label2]{Yonghyun Song}
\author[label3]{Hyun-Jic Oh}
\author[label4,label4.1]{Dalit Ben-Yosef}
\author[label2,label5]{Daniel Needleman}
\author[label1]{Hanspeter Pfister}
\fntext[1]{* These authors contributed equally to this work.}

\affiliation[label1]{organization={School of Engineering and Applied Sciences, Harvard University},
            city={Allston},
            state={MA},
            country={USA}}

\affiliation[label2]{organization={Department of Molecular and Cellular Biology, Harvard University},
            city={Cambridge},
            state={MA},
            country={USA}}

\affiliation[label3]{organization={College of Informatics and Department of Computer Science and Engineering, \\Korea University},
            city={Seoul}, 
            country={South Korea}}

\affiliation[label4]{organization={Fertility \& IVF Institute, Tel Aviv Sourasky Medical Center},
            city={Tel Aviv},
            country={Israel}}

\affiliation[label4.1]{organization={CORAL - Center Of Regeneration And Longevity, Tel Aviv University},
            city={Tel Aviv},
            country={Israel}}

\affiliation[label5]{organization={Center for Computational Biology, Flatiron Institute},
            city={NY},
            state={NY},
            country={USA}}

\begin{abstract}
Automating embryo viability prediction of in vitro fertilization (IVF) is a significant yet challenging task due to the limited availability of labeled pregnancy outcome data—only a small fraction of embryos transferred back to the uterus are labeled with associated outcomes. Self-supervised learning (SSL) offers a solution by enabling the use of both labeled and unlabeled data to improve downstream task performance. However, existing SSL methods for videos cannot be directly applied to the representation learning of embryo development videos due to two main challenges. Firstly, embryo time-lapse videos often consist of hundreds of frames, demanding significant GPU memory when using conventional SSL methods. Secondly, our dataset consists of videos with varying lengths and numerous outlier frames, causing traditional SSL methods based on video alignment to struggle with substantial temporal semantic misalignments between videos. To address these two challenges, we introduce the Spatial-Temporal Pre-Training (STPT) method. To make representation learning more memory-efficient, STPT comprises two separate encoder training stages: the spatial stage and the temporal stage. In each stage, only one encoder is trained while the other remains frozen. To mitigate the impact of temporal semantic misalignment, STPT avoids frame-by-frame alignment learning among different videos. Instead, the first stage focuses on learning from alignments within each video and its temporally consistent augmentations. After obtaining the embedding for each video, the second stage then learns the relationships between the embeddings of different videos. Our approach effectively manages the computational demands of lengthy videos and the varied temporal semantics of embryo development. We conducted experiments on 23,027 time-lapse videos, of which 3,286 were labeled. Compared to baselines, STPT achieves the highest AUC of 0.635 (95\% CI: 0.632-0.638) while utilizing limited computational resources.
\end{abstract}

\begin{graphicalabstract}
\includegraphics[width=\textwidth]{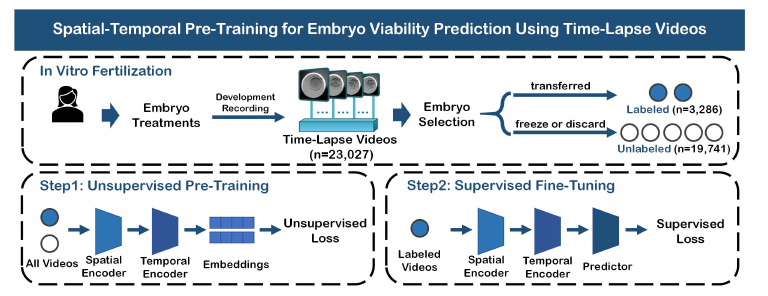}
\end{graphicalabstract}

\begin{keyword}
in vitro fertilization \sep self-supervised learning \sep time-lapse video \sep embryo viability prediction
\end{keyword}

\end{frontmatter}


\section{Introduction}
\label{sec:introduction}

In vitro fertilization (IVF) represents a significant advancement in reproductive technology, offering hope to millions of couples facing infertility worldwide~\cite{wang2021ivf,kwan2021monitoring,van2010predictive}. 
IVF involves stimulating patients to produce multiple oocytes (eggs), which are subsequently retrieved and fertilized in vitro. The resulting embryos are cultured for 2-5 days, with selected embryos transferred to the maternal uterus to initiate pregnancy, while surplus viable embryos are cryopreserved for future use. Although transferring multiple embryos can enhance the probability of conception, it concurrently increases the risk of multiple pregnancies. These pregnancies are associated with higher rates of maternal and neonatal morbidity and mortality~\cite{norwitz2005maternal}. Therefore, limiting embryo transfer to a single, optimally selected one is imperative to maximize the likelihood of a healthy singleton birth~\cite{lee2016elective}, which remains challenging~\cite{racowsky2011national}.

The traditional approach to embryo selection relies on morphological analysis of microscopic imaging. After fertilization, embryos undergo various developmental stages, from pronuclei alignment to blastocyst formation, which provides rich features indicative of embryo viability. Traditionally, clinicians have assessed embryos by manually examining characteristics at these different stages, such as cell number, cell shape, cell symmetry, the presence of cell fragments, and the appearance of the blastocyst~\cite{elder2000vitro}. In recent years, many clinicians have started using time-lapse microscopy incubators that continuously record the development of embryos without disrupting their culture conditions~\cite{armstrong2019time,khan2016deep,lau2019embryo}. Despite this technological progress, analyzing these videos remains a manual process, which is both labor-intensive and subjective~\cite{lukyanenko2021developmental,jang2023amodal}. 

Recent studies have employed deep learning methods to automate the analysis of time-lapse videos of embryos~\cite{supervised1,supervised2,supervised3,wang2024generalized}. In this context, predicting embryo viability is considered a supervised learning problem, where models are trained using ground truth data on pregnancy outcomes. However, a significant challenge arises since only a small fraction of all embryos are transferred back to the uterus, and not all fertility clinics consistently track pregnancy outcomes. Consequently, embryo datasets often suffer from a scarcity of labeled samples. In our dataset, for instance, only 13.8\% of the embryos were transferred and thus had associated outcome labels.

Self-supervised learning (SSL) methods~\cite{simclr,mae,simsiam,infonce} have emerged as a powerful pre-training strategy, particularly useful in situations with scarce labeled data. These methods enable models to learn representations from unlabeled data during the pre-training phase, without the need for explicit supervision. This process helps the model better understand the distribution and characteristics of the entire dataset, thereby enhancing its performance on downstream tasks. Although numerous SSL methods have been developed for videos~\cite{videomae,CARL,CVRL,tmi,align,dtw}, they are not fully compatible with our specific scenario due to two main challenges.

\textbf{GPU Memory Requirements.} 
Previous studies have shown that larger batch sizes can improve the effectiveness of SSL methods for image representation learning~\cite{moco, debiased}. However, a video may have hundreds of frames, resulting in significant GPU memory requirements. Consequently, most self-supervised learning algorithms for videos, such as VideoMAE~\cite{videomae} and VCLR~\cite{vclr}, are primarily designed for short videos, \ie, less than 32 frames. Moreover, these methods still demand substantial computational resources, \eg, 64 Tesla V100 GPUs for VideoMAE using ViT as a backbone~\cite{videomae}. In contrast, most videos in our dataset contain 100 to 500 frames, requiring the computational load to be exceedingly high even when frame sampling is employed.

\begin{figure*}[t]
\centering
\includegraphics[width=1.0\linewidth, height=0.1136\linewidth]{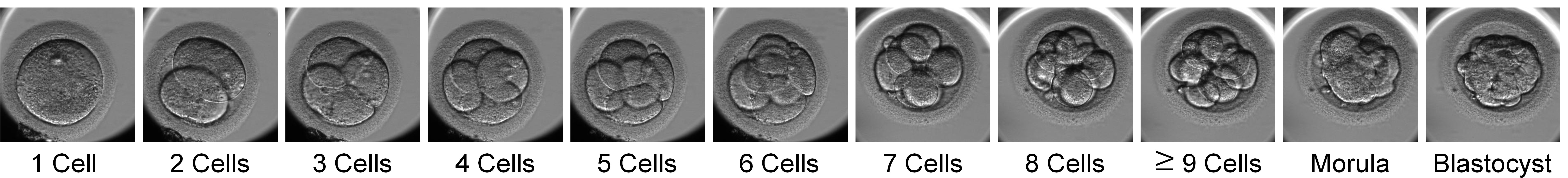}
\caption{Examples of developmental stages from 1-cell to blastocyst. Among these stages, 9 classes are for cleavage-stage embryos (one for each of 1–8 cells and one for $\ge$ 9 cells), and one class each for morula (M) and blastocyst (B).}
\label{fig:stages}
\end{figure*}

\begin{figure*}[t]
\centering
\includegraphics[width=1.0\linewidth,height=0.2657\linewidth]{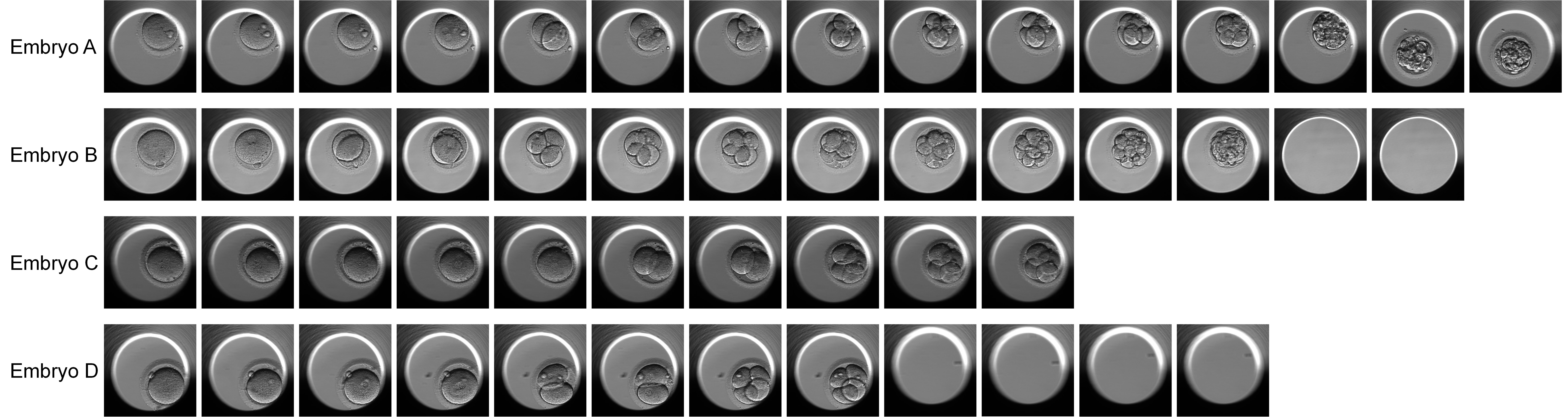}
\caption{Examples of embryo developments: Embryo A and Embryo B progress from a single cell to a blastocyst, with Embryo B developing faster and being taken out earlier, leaving an empty well. In contrast, Embryo C and Embryo D have relatively short videos, reaching only the 4-cell stage, with Embryo D being taken out earlier. The length of videos varies due to different end-stage developments, but they also vary even when end-stage developments are the same due to different development speeds. Additionally, videos may contain redundant frames that capture an empty well even after removing embryos.}
\label{fig:embryo}
\end{figure*}

\begin{figure*}[t]
\centering
\includegraphics[width=\linewidth]{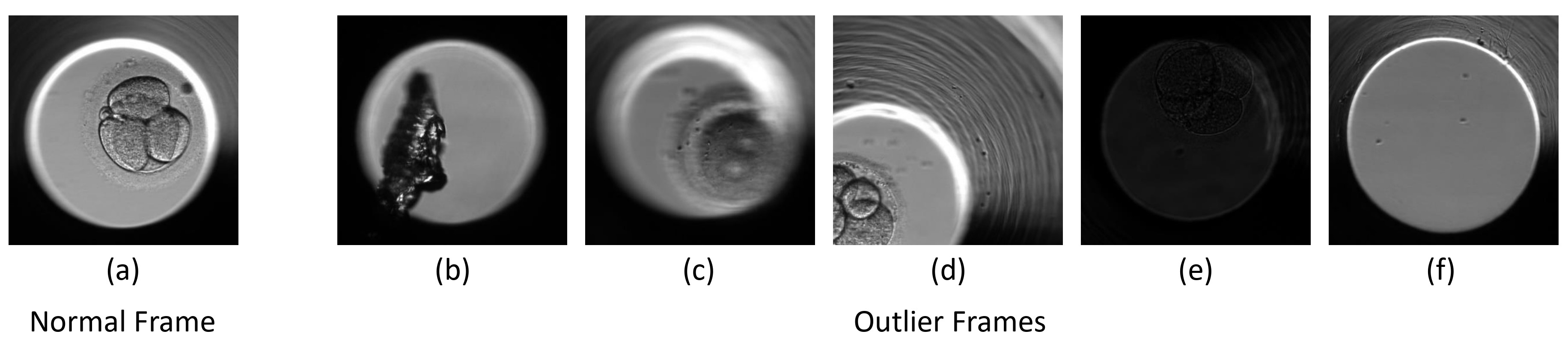}
\caption{Examples of the normal frame and the outlier frames. (a) Normal frame. (b) Dirt or a piece of glass pipette left in the well. (c) Out of focus. (d) Center misalignment. (e) Low exposure. (f) Empty well.}
\label{fig:outlier}
\end{figure*}

\textbf{Temporal Semantic Misalignment.} 
To apply self-supervised learning for long videos (with more than 128 frames), several approaches~\cite{tcc,tmi,align,dtw} have employed temporal alignment across different samples to learn frame representations and relationships. However, these methods require videos with similar semantics that are temporally aligned. Therefore, self-supervised learning using temporal alignment necessitates datasets with videos of the same events that can be aligned. This assumption does not hold in our dataset. In clinical IVF, embryos are cultured for different durations depending on various factors, such as embryo condition or development speed. Some embryos are transferred around day 3 at the cleavage stage, while others are transferred around day 5 at the blastocyst stage. Some embryos may degenerate during development and are thus removed at early stages.
The example of each development stage is shown in Fig.~\ref{fig:stages}.
This variability not only results in different video lengths but also leads to significant differences in temporal semantic information due to the varying developmental stages, as illustrated in Fig.~\ref{fig:embryo}.
Moreover, temporal misalignment across different videos may occur due to outlier frames with various causes, as shown in Fig.~\ref{fig:outlier}. We observe several outlier frames caused by debris in the well, out-of-focus imaging, misalignment of the microscopy imaging center, low exposure, or empty wells.

To address these challenges, we propose Spatial-Temporal Pre-Training (STPT) with an augmentation-based self-alignment scheme. One significant challenge in utilizing embryo videos for self-supervised learning is their considerable length, typically ranging from 100 to 500 frames. Directly applying video self-supervised learning methods~\cite{videomae,vclr} is impractical due to excessive GPU memory requirements. To mitigate this issue, we decouple spatial and temporal learning into separate training stages. The spatial stage is designed to train a spatial encoder to learn spatial embeddings. The temporal stage focuses on learning temporal relations to integrate these frame embeddings into a video embedding. This decoupled training approach significantly reduces computational burdens since spatial and temporal relations are trained separately.

Another challenge is the temporal misalignment among videos. While alignment methods have proven effective for embryo viability prediction~\cite{tmi}, they are only applicable when a dataset contains videos with temporally aligned semantic information. However, embryos develop at different speeds, resulting in videos with varying developmental stages and durations, as shown in Fig.~\ref{fig:embryo}. Furthermore, outlier frames can introduce noise in alignment-based methods. Collecting a temporally aligned training set requires extensive data inspection and filtering, which is costly and reduces the number of usable samples which is an important consideration given the expense of medical data collection. To circumvent this, we propose aligning two different augmented versions of the same video, ensuring temporal semantic alignment as both versions originate from the same video. This augmentation and alignment method is robust to variations in video length and the presence of outlier frames.

More specifically, in the spatial stage, we propose self-cycle temporal alignment to train the spatial encoder by alignment between two differently augmented versions of the same video. In the temporal stage, each frame is encoded into a frame embedding using a spatial encoder. We then use a temporal encoder to encode a video embedding from these frame embeddings. A contrastive loss~\cite{simclr} is applied to the video embeddings across all the unlabeled data, enabling the temporal encoder to learn video semantic information.

After the two-stage pre-training, we attach a Multilayer Perceptron (MLP) head to the output of the temporal encoder for fine-tuning the embryo viability prediction task. This MLP head is trained on the labeled data to obtain the final prediction. Our two-stage method significantly reduces the required computational resources as only one module is trained at each stage while the other modules are frozen. To validate the effectiveness of our method, we conduct five-fold cross-validation on labeled embryos. Experimental results demonstrate the effectiveness of our method over competing approaches in both performance and computational efficiency.

\section{Related Work}
\label{sec:Related_Work}

Traditionally, clinicians assess embryos by visually inspecting morphokinetic traits such as cell number, shape, symmetry, the presence of fragments, and blastocyst appearance~\cite{elder2000vitro}, a process that is both time-consuming and prone to subjective interpretation. To address these challenges, recent studies~\cite{supervised1,supervised2,supervised3,wang2024generalized} have incorporated deep learning techniques to automate the analysis of time-lapse videos. The task of predicting embryo viability can be formulated as a supervised learning problem guided by pregnancy outcomes. However, in practice, most embryos do not have direct outcome labels, limiting the applicability of supervised learning-based methods.

Self-supervised learning (SSL), often used during the pre-training phase for representation learning, enables the use of both labeled and unlabeled data to learn feature representations without explicit supervision. This pre-training process learns the representation of the entire data distribution, improving performance when fine-tuned on downstream tasks. Various representation learning approaches have been proposed for images, with two of the most prevalent being contrastive learning-based and masked autoencoder-based methods. SimCLR~\cite{simclr} is a contrastive learning-based SSL method that learns visual representations by maximizing the agreement between differently augmented views of the same image. MAE~\cite{mae}, on the other hand, is a reconstruction-based SSL approach that focuses on masking parts of the input image and training the model to reconstruct the missing portions, thus capturing comprehensive feature representations.

Image-based SSL methods can still be applied to video-based representation learning. VideoMAE~\cite{videomae} is a video-based adaptation of MAE~\cite{mae}, focusing on temporal and spatial masking to effectively learn video representations by reconstructing missing segments. VCLR~\cite{vclr} leverages the temporal coherence of videos to learn robust video representations through the comparison of different clips from the same video. CVRL~\cite{CVRL} applies the SimCLR~\cite{simclr}  to video representation learning by constructing temporally-consistent spatial augmentations. However, these methods are mainly designed for short videos, usually less than 32 frames, and require significant computational resources. 
Temporal cycle consistency (TCC)~\cite{tcc} proposes to align videos containing sequential processes of the same events carried out in the same order. However, this method~\cite{tcc,tmi} requires videos with similar semantics that are temporally aligned. Nevertheless, the videos in our dataset contain different developmental stages and thus can not be aligned with each other. To address this limitation, CARL~\cite{CARL} proposes aligning the augmented views of the same video. Yet, CARL employs random sampling for frames, resulting in suboptimal performance on our dataset (as shown in Table \ref{tab:augmentation}). This may be because random sampling changes the relative speeds of different developmental stages, which are crucial characteristics of embryo viability.

\section{Method}
\label{sec:method}

\begin{figure*}[t]
\centering
\includegraphics[width=\linewidth,height=0.5168\linewidth]{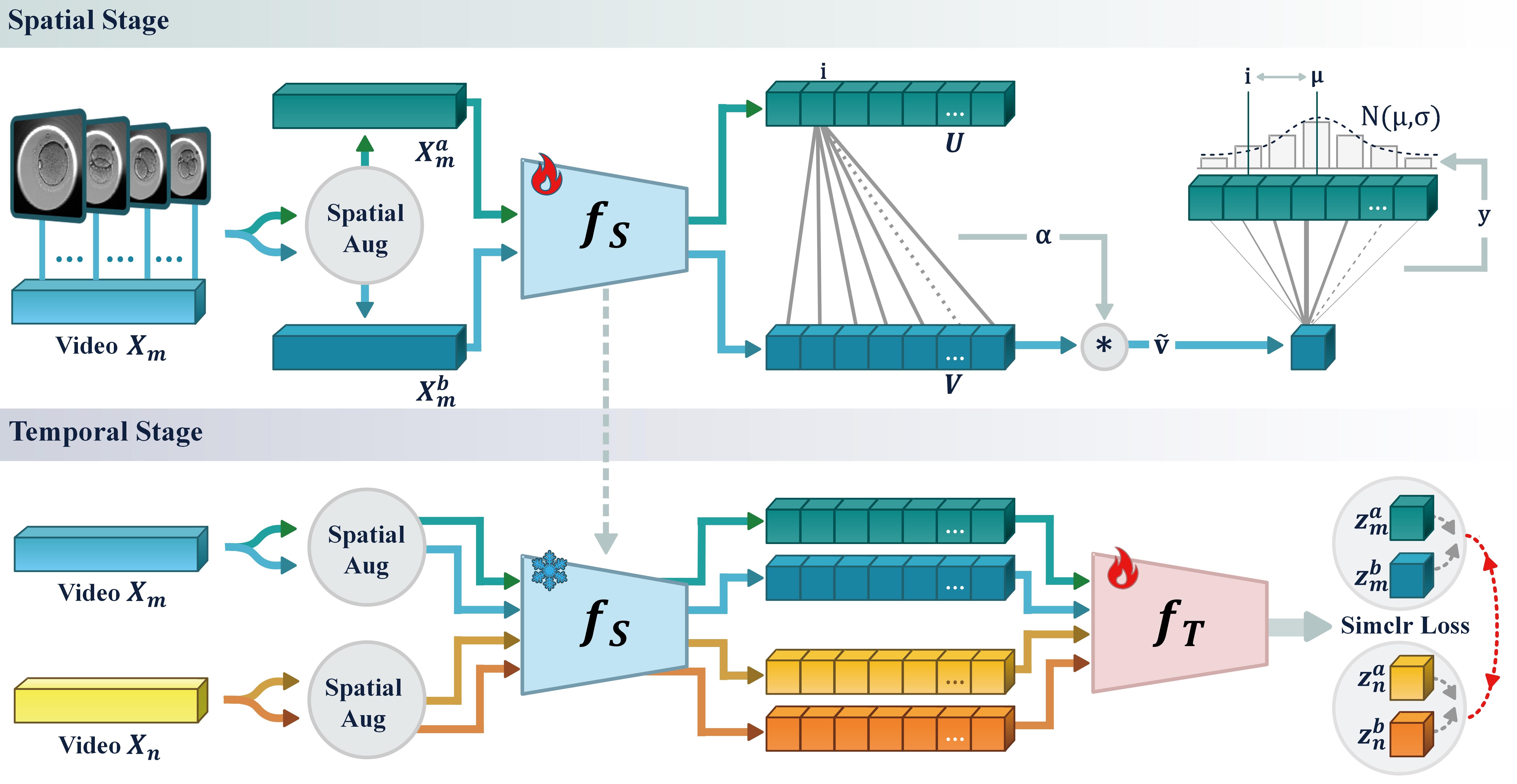}
\caption{The overview of our two-stage pre-training method. Each cube represents an embedding per frame encoded by the spatial encoder. A trainable module is indicated by a fire icon, while a frozen module is indicated by a snowflake icon.}
\label{fig1}
\end{figure*}

In this section, we introduce Spatial-Temporal Pre-Training (STPT), which includes two pre-training stages: the spatial stage and the temporal stage, as demonstrated in Fig~\ref{fig1}. In the spatial stage, we use self-cycle temporal consistency loss to learn the frame representation. In the temporal stage, frame embeddings are input into the temporal encoder to obtain video embedding. This two-stage pre-training method separates the learning of spatial and temporal encoders, thereby enhancing the efficiency of the training process.

\subsection{Spatial Stage}

Given the m-th input video $X_m$ with \(T\) frames, \(X_m = \{x_{m,i} | 1\leq i \leq T\}\), we construct two augmented views \(X_m^a = \{x_{m,i}^a | 1\leq i \leq T\}\) and \(X_m^b = \{x_{b,i} | 1\leq i \leq T\}\) independently through a series of temporally-consistent spatial augmentations. ``temporally-consistent'' means that the same augmentation is applied to all frames within the same video.

To extract spatial features, the spatial encoder \(f_S\) embeds each frame \(x_{m,i}\) to a vector.
Let the embeddings of two augmented videos, $X_m^a$ and $X_m^b$, extracted by frame encoder $f_S$ be denoted as \(U = \{u_i | 1\leq i \leq T\}\) and \(V = \{v_i | 1\leq i \leq T\}\) respectively.
The spatial encoder \(f_S\) is trained to align the embeddings of two augmented videos $U$ and $V$ by temporal cycle consistency (TCC)~\cite{tcc,tmi}.

The alignment of two augmented videos can be achieved by enforcing TCC between them. TCC enforces each point in one video finds its nearest neighbor in the other video and vice versa, maintaining a consistent cycle.
Let \( u_i\) be a frame in $U$. We first determine its nearest neighbor in \( V \) as $v_j$, and the nearest neighbor of $v_j$ back in $U$ as $u_k$, as follows:
\begin{equation}
    v_j = \arg \min_{v \in V} \| u_i - v \|, \quad u_k = \arg \min_{u \in U} \| v_j - u \|.
\end{equation}
The embedding \( u_i \) is cycle-consistent if and only if \( i = k \). However, this is not a differentiable process.
TCC introduces the concept of a soft nearest neighbor to make the process differentiable.
For each point \( u_i \), the soft nearest neighbor \( \tilde{v} \) in \( V \) is defined using a softmax function over the distances:
\begin{equation}
    \tilde{v} = \sum_{j=1}^{T} \alpha_j v_j, \quad \alpha_j = \frac{e^{-||u_i - v_j||^2}}{\sum_{t=1}^{T} e^{-||u_i - v_t||^2}}.
\end{equation}
Here, \( \alpha \) represents the similarity distribution between \( u_i \) and each \( v_j \in V \).
Then, the nearest neighbor \( u_k \) of this soft nearest neighbor \( \tilde{v} \) back in \( U \) is searched. To enforce cycle consistency, each frame index is treated as a separate class. Then, cycle consistency becomes a classification problem, where the task is to identify the index that satisfies cycle consistency. 
The TCC classification loss is defined as follows:
\begin{equation}
    \mathcal{L}_{cls} = -\sum_{x_i \in X}\sum_{k=1}^{T} y_k \log(\hat{y}_k), \quad \hat{y}_k = \frac{e^{-||\tilde{v} - u_k||^2}}{\sum_{t=1}^{T} e^{-||\tilde{v} - u_t||^2}}
    \label{eq:loss_cls}
\end{equation}
where $y$ is a one-hot vector at index $i$, \ie, $y_i=1$. Therefore, $\mathcal{L}_{cls}$ is minimized when $i = \arg\max_{k} \hat{y}_k$.

Although this classification loss defines a differentiable cycle-consistency loss function, it does not consider the temporal proximity of the points. To enforce temporal proximity, TCC penalizes the model less if the frame cycles back to closer neighbors, as opposed to frames that are farther away in time.
The variable $\hat{y}$ in Eq.~\ref{eq:loss_cls} represents a discrete distribution of similarities, expected to peak around the $i^{th}$ index.
To ensure this, TCC apply a Gaussian prior to \( \hat{y} \) by minimizing the normalized squared deviation \(\frac{|i - \mu|^2}{\sigma^2}\) and add variance regularization to keep \( \hat{y} \) concentrated around \( i \).
The TCC regression loss per video is defined as:
\begin{equation}
\label{eq:tcc}
\mathcal{L}_{TCC} = \sum_{x_i \in X}\frac{|i - \mu|^2}{\sigma^2} + \lambda \log(\sigma)
\end{equation}
where \(\mu = \sum_{k=1}^{N} \hat{y}_k \cdot k \) and \(\sigma^2 = \sum_{k=1}^{N} \hat{y}_k \cdot (k - \mu)^2 \), with \(\lambda\) being the regularization weight.

This TCC regression loss $\mathcal{L}_{TCC}$ can be viewed as an enhanced version of classification loss $\mathcal{L}_{cls}$, incorporating the information of temporal proximity. As shown in Fig.~\ref{fig1}, in the spatial stage, we apply this TCC regression loss to align the representations of two augmented views generated from the same video. The overall process of the spatial pre-training is illustrated in Alg.~\ref{alg:spatial}.

\begin{algorithm}[t]
\setstretch{1.2} 
\caption{Spatial Stage Pre-training}
\label{alg:spatial}
\begin{algorithmic}
\State \textbf{Input\hspace{0.72em}:} $f_S$: spatial encoder, $\mathcal{D}=\{X_m\}_{m=1}^N$: training set
\State \textbf{Output:} Updated $f_S$
\For{$X_m$ \textbf{in} $\mathcal{D}$} \Comment{Sample a video}
    \State $X_m^a,\ X_m^b \ \gets \ aug(X_m),\ aug(X_m)$
    \State $U,\ V \gets f_S(X_m^a),\ f_S(X_m^b)$ 

    \State $L \gets L_{TCC}(U, V)$ \Comment{Eq.~\ref{eq:tcc}}

    \State $L.\text{backward}()$ \Comment{Back-propagate}
    \State update($f_S$) \Comment{Adam update for $f_S$}
\EndFor
\end{algorithmic}
\end{algorithm}

\begin{algorithm}[t]
\setstretch{1.2} 
\caption{Temporal Stage Pre-training}
\label{alg:temporal}
\begin{algorithmic}
\State \textbf{Input\hspace{0.72em}:} $f_S$: spatial encoder, $f_T$: temporal encoder, \\\hspace{3.75em}$\mathcal{D}=\{X_m\}_{m=1}^N$: training set
\State \textbf{Output:} Updated $f_T$
\For{$\mathcal{B}$ \textbf{in} $\mathcal{D}$} \Comment{Sample a mini-batch}
    \State $\mathcal{B}^a, \ \mathcal{B}^b \ \gets \ \mathrm{aug}(\mathcal{B}), \ \mathrm{aug}(\mathcal{B})$ 
    \State \textbf{with} no\_grad(): \Comment{Freeze $f_S$}
    \State \hspace{1.5em}$\mathcal{U}, \ \mathcal{V} \ \gets \ f_S(\mathcal{B}^a), \ f_S(\mathcal{B}^b)$ 
    
    \State $Z^a, \ Z^b \ \gets \ f_T(\mathcal{U}), \ f_T(\mathcal{V})$ 
    \State $L \gets L_{con}(Z^a, Z^b)$ \Comment{Eq.~\ref{eq:contrastive}}
    \State $L.\text{backward}()$ \Comment{Back-propagate}
    \State update($f_T$) \Comment{Adam update for $f_T$}
\EndFor
\end{algorithmic}
\end{algorithm}

\subsection{Temporal Stage}

In this stage, the spatial encoder $f_S$ is frozen, and the temporal encoder $f_T$ is trained by contrastive loss~\cite{simclr}. 
The temporal encoder inputs all frame embeddings of a video and then outputs a video embedding. 
Suppose we sample $n$ videos to construct a batch $\mathcal{B}$ and augment the batch to make $\mathcal{B}^a$ and $\mathcal{B}^b$.
Then we get two batches of frame representations $\mathcal{U}=f_S(\mathcal{B}^a)$ and $\mathcal{V}=f_S(\mathcal{B}^b)$ using spatial encoder.
Let's denote $z_m^a=f_T(U)$ and $z_m^b=f_T(V)$ as the encoded representations of the two augmented views of the $m$-th input video, where $U \in \mathcal{U}$ and $V \in \mathcal{V}$. Then, the contrastive loss is defined as:

%
\begin{equation}
\label{eq:contrastive}
\mathcal{L}_{con} = \frac{1}{n}\sum^n_{m=1} -\log \frac{\exp(\text{sim}(z_m^a, z_m^b) / \tau)}{
\sum_{k=1}^{2n} 
\textbf{1}_{[k \neq m]} \exp(\text{sim}(z_m, z_k) / \tau)}
\end{equation}

where $z_k \in \{\mathcal{U}, \mathcal{V}\}$, $\text{sim}(a, b)$ is the cosine similarity between two $\ell_2$ normalized vectors \(a\) and \(b\), \(\textbf{1}_{[\cdot]}\) is an indicator excluding the self-similarity of the encoded video \(z_i\) from the denominator, 
and \( \tau \) is a temperature parameter. 
Following common practice, we add the sine-cosine positional encoding on top of the input embeddings to encode the temporal information.
The overall process of the temporal pre-training is illustrated in Alg.~\ref{alg:temporal}.

\subsection{Fine-tuning Stage}

In this stage, we adopt labeled data to train the model for embryo viability prediction. For each patient with pregnancy outcomes, $\mathrm{n\_transferred}$ is the number of embryos transferred and $\mathrm{n\_births}$ is the number of births. Our goal is to predict embryo viability formulated as 
$p = \frac{\mathrm{n\_births}}{\mathrm{n\_transferred}}$, which is defined as the number of births over the number of embryos transferred.
For the final prediction, we attach a classifier $f_C$ to the output of the temporal encoder. $f_C$ is MLP, which consists of two fully connected layers with ReLU activation in between.
Let's denote $z_m$ as the video embedding of the labeled video $X_m$ and $\hat{p}_m = f_C(z_m)$ as the prediction for viability.
For supervised training, we employ the Huber loss~\cite{huber}, which is defined for each video $X_m$ as follows:

\begin{equation}
L_{huber}
=
\begin{cases} 
\frac{1}{2} (p_m - \hat{p}_m)^2 & \text{for } |p_m - \hat{p}_m| \leq \delta, \\
\delta \cdot (|p_m - \hat{p}_m| - \frac{1}{2} \delta) & \text{otherwise}.
\end{cases}
\end{equation}
where $\delta$ is a hyperparameter, with a default value of 0.2. To conserve computational resources and mitigate overfitting, we keep the spatial encoder frozen during the fine-tuning stage. We train both the temporal encoder and the classifier together, as the temporal encoder is relatively lightweight compared to the spatial encoder.

\section{Experiments}
\label{sec:exp}

\subsection{Experiment Setup}

\noindent\textbf{Dataset.}
We gathered data from 3,695 IVF treatment cycles with 23,027 embryos imaged every 20 min up to the first 5 days of development where each image size is $500\times500$ pixels. This corresponds to approximately 6 million images of embryos. Among all the data, 1700 treatment cycles with 3286 embryos are transferred with exact pregnancy outcomes. Out of 1700 treatments, 260 treatments are successful with equal or more than one live birth. It's important to note that each treatment cycle fertilizes multiple embryos, and only qualified embryos are selected for transfer. Additionally, some cycles freeze all embryos for future use instead of immediate transfer. Consequently, the proportion of embryos with treatment outcomes is relatively small compared to the volume of raw data collected. 

\begin{figure*}[t]
\centering
\includegraphics[width=\linewidth,height=0.3494\linewidth]{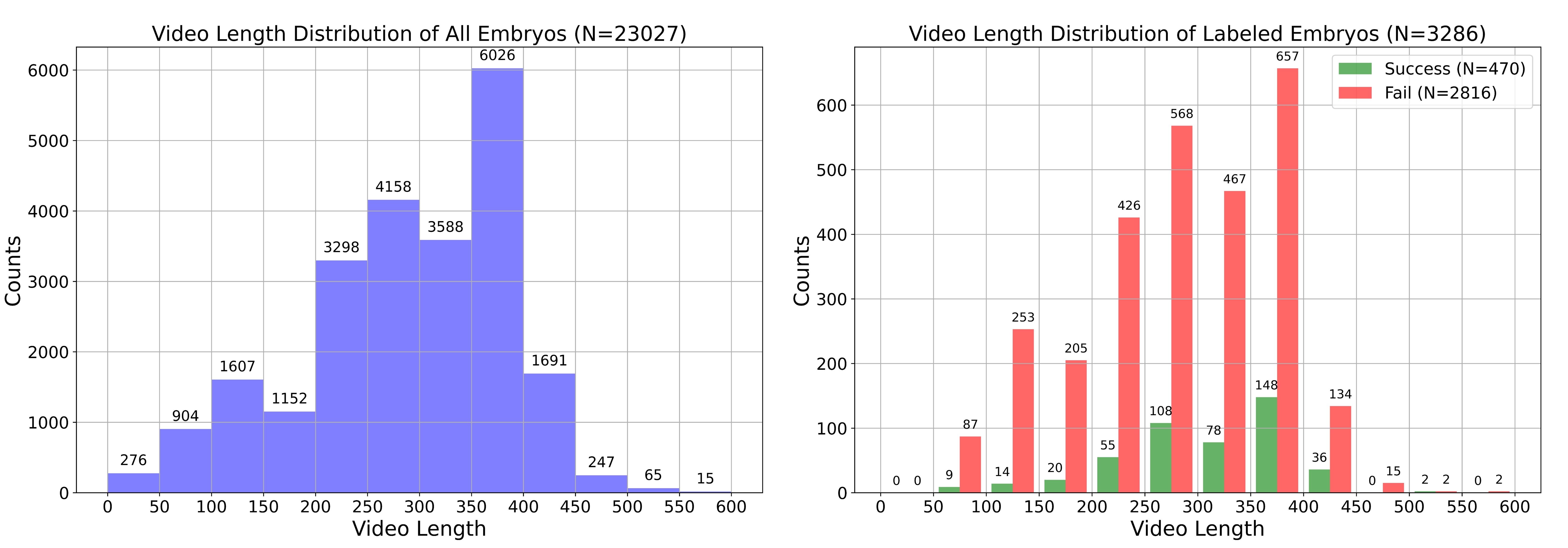}
\caption{Distribution of video lengths. N represents the total number of videos. (a) demonstrates the distribution of all videos. (b) demonstrates the distribution of labeled videos; for each interval, the left column represents the number of videos with successful outcomes, while the right column represents the number of videos with failed outcomes.}
\label{fig:length}
\end{figure*}

The video length distributions are shown in Fig.~\ref{fig:length}, where (a) demonstrates the distribution of all videos while (b) demonstrates the distribution of labeled videos. According to this figure, the length distribution of labeled embryos is similar to that of all embryos. Additionally, the length distribution of successful embryos is comparable to that of failed embryos. Overall, video length does not clearly distinguish between labeled and unlabeled embryos or successful and failed embryos. 

\noindent\textbf{Preprocessing.}
We have identified several outlier frames in the embryo videos, typically associated with anomalies in laboratory conditions, as illustrated in Fig.~\ref{fig:outlier}. To remove these frames, we employ Canny Edge Detection (CED)~\cite{canny} as an initial data cleansing step. Image gradients are calculated for each frame, and outlier frames caused by blur, empty wells, or light saturation are identified by their low gradient values and removed from the videos.

\noindent\textbf{Evaluation.}
For evaluation, we use the area under the receiver operating characteristic curve (AUROC)~\cite{roc}. In the embryo viability prediction task, we set the ground truth label to `1' for all embryos transferred together (instead of $\mathrm{\frac{n_{births}}{n_{transferred}}}$) if the treatment is successful when computing AUROC. 
To ensure the statistical significance and robustness of our experiments, we split our dataset into 5 folds and adopt cross-validation. For each fold, we conduct the experiment three times and report the results with 95\% confidence intervals.

\subsection{Implementation details}

\noindent\textbf{Spatial stage.} 

For the spatial encoder, we adopt DeiT-Tiny~\cite{deit} and start SSL training from pre-trained weights. Input frames are resized from 500 × 500 to 224 × 224. To enable memory-efficient training, we sub-sample every 4 frames, resulting in 90 frames per video. This covers 360 frames in the original video. The batch size is set to 4 and the number of augmented views is set to 4. The learning rate is set to 1e-5 and the weight decay is set to 1e-4. The total training epoch is set to 50, which is sufficient to ensure convergence for all methods used in the experiments.

\noindent\textbf{Temporal stage.} For the temporal encoder, we use 4 transformer layers with hidden size 192 and 8 heads to model temporal context. Videos are clipped to have a maximum of 360 frames since this corresponds to the first 5 days of observation, where each frame is captured at 20-minute intervals. To enable memory-efficient training, we sub-sample every 4 frames, resulting in 90 frames per video. The batch size is set to 128 and the number of augmented views for each video is set to 2. The learning rate is set to 1e-5, and the weight decay is set to 1e-5. We set \( \tau \) to 0.07, which is the default value in many contrastive learning studies~\cite{kukleva2023temperature}.

\noindent\textbf{Fine-tuning stage.} The MLP size is set to 192. For a fair comparison, we use the same fine-tuning parameters and loss function across all methods. Specifically, we adopt the Huber loss~\cite{huber} and set \(\delta\) in the loss to 0.2. The batch size is set to 2. The learning rate is set to \(1 \times 10^{-4}\) and the weight decay is set to \(1 \times 10^{-5}\). Each model is trained for a total of 10 epochs.

\noindent\textbf{Augmentation.}
Following CVRL~\cite{CVRL}, for all training stages, we adopt temporally-consistent spatial augmentations including applying random flip, rotation, noise, brightness, and contrast jitter. For all experiments, we use the Adam optimizer and one A100 GPU.

\subsection{Results}

\noindent\textbf{Performance.}

\begin{table*}[t!]
\renewcommand\arraystretch{1.5}
\centering
\caption{Confidence interval of AUCROC for different SSL pre-training methods on 5-fold cross-validation}
\resizebox{\linewidth}{!}{
\setlength{\tabcolsep}{10pt}
\begin{tabular}{lcccccc} 
\hline
Methods   & Fold1        & Fold2   & Fold3    & Fold4     & Fold5    & Mean       \\ 
\hline
Deit~\cite{deit} & \(0.554 \pm 0.013 \) & \(0.573 \pm 0.007 \) & \(0.573 \pm 0.013 \) & \(0.472 \pm 0.033 \) & \(0.633 \pm 0.006 \) & \(0.561 \pm 0.005 \)  \\ 
\hline
TCC~\cite{tcc}               & \(0.586 \pm 0.004 \) & \(0.573 \pm 0.006 \) & \(0.616 \pm 0.016 \) & \(0.506 \pm 0.014 \) & \(0.705 \pm 0.003 \) & \(0.597 \pm 0.003 \)  \\
CARL~\cite{CARL}              & \(0.574 \pm 0.003 \) & \(0.580 \pm 0.007 \) & \(\mathbf{0.632 \pm 0.004} \) & \(0.499 \pm 0.013 \) & \(0.704 \pm 0.006 \) & \(0.598 \pm 0.003 \)  \\
CVRL~\cite{CVRL}              & \(0.598 \pm 0.006 \) & \(0.573 \pm 0.001 \) & \(0.627 \pm 0.004 \) & \(0.526 \pm 0.004 \) & \(0.694 \pm 0.001 \) & \(0.604 \pm 0.002 \)  \\
Ours      & \(\mathbf{0.605 \pm 0.010} \) & \(\mathbf{0.618 \pm 0.000} \) & \(0.630 \pm 0.008 \)  & \(\mathbf{0.619 \pm 0.009} \) & \(\mathbf{0.705 \pm 0.000 }\) & \(\mathbf{0.635 \pm 0.003} \)  \\
\hline
\end{tabular}
}
\label{tab:results}
\end{table*}

We compare different self-supervised learning (SSL) pre-training methods for videos on our dataset. To ensure the robustness of our method with statistical significance, we employ 5-fold cross-validation on all labeled data and repeated the experiments three times for each fold. The results are shown in Table~\ref{tab:results}. Each method in the table is used for pre-training before supervised fine-tuning. For Deit~\cite{deit}, we do not apply any additional pre-training on the unlabeled embryo videos but directly use the pre-trained representation from natural images. The other methods in the table include pre-training on embryo videos followed by fine-tuning.

The results demonstrate the importance of pre-training on unlabeled embryo videos, as all pre-training methods outperformed Deit, which does not apply pre-training. Although the fine-tuning stage involves supervised learning, the results indicate that the limited amount of labeled data is insufficient for training a robust model. Among the pre-training methods, our proposed method achieves the highest AUC of $0.635 \pm 0.003$  (p value $<$ 0.0001). We conjecture that the performance gap between the proposed method and the competing methods is due to our self-augment and alignment technique, which is robust to videos with temporal misalignment and outlier frames that may otherwise deteriorate the training process.

\begin{table}[t]
\centering
    \caption{GPU memory requirement of pre-training when using different batch sizes}
\centering
\resizebox{0.7\linewidth}{!}{
\begin{tabular}{ccccccc} 
\hline
Batch Size                 & 4     & 6     & 8     & 10    & 12 & 14    \\ 
\hline
CVRL~\cite{CVRL} (GB)         & 22.19 & 32.81 & 43.55 & 54.20 & 64.84  & 75.58\\
Ours (GB)             & 3.21  & 4.52  & 5.83  & 7.11  & 8.40 & 9.71 \\ 
\hline
ratio $( \displaystyle \frac{\mathrm{CVRL}}{\mathrm{Ours}})$  & 6.91  & 7.26  & 7.47  & 7.62  & 7.72 & 7.78 \\
\hline
\end{tabular}
}
\label{tab:GPUs}
\end{table}

\noindent\textbf{Computational resource.} Typically, pre-training methods train the spatial encoder and the temporal encoder simultaneously in a single stage. However, our approach separates the training into two stages: when one module is trained, the other modules are frozen. To demonstrate that our two-stage method is more memory-efficient than the combined stage method, we compare the GPU memory required for training each method at different batch sizes, as shown in Table~\ref{tab:GPUs}. The table shows that, for the same batch sizes, the GPU memory required for our method is significantly smaller than that of CVRL~\cite{CVRL}, which trains both the spatial and temporal encoders simultaneously in an end-to-end fashion.

We compare the available batch sizes when these two methods use the same GPU memory. When the GPU memory is limited to 64GB, the maximum batch size for CVRL method is 12, whereas our method can accommodate a batch size of up to 128. The results in Table~\ref{tab:64GB} show that the larger batch size helps improve the effectiveness of contrastive learning, which is consistent with the findings from the previous works~\cite{moco,debiased}.

\begin{table*}[t]
\centering
    \caption{Ablation on batch sizes when GPU memory is limited to 64GB}
\renewcommand\arraystretch{1.5}
\resizebox{\linewidth}{!}{
\begin{tabular}{cccccccc}
\hline
      Methods        & Batch Size & Fold1        & Fold2   & Fold3    & Fold4     & Fold5    & Mean       \\ 
\hline
w/o pre-training & ----  & \(0.554 \pm 0.013 \) & \(0.573 \pm 0.007 \) & \(0.573 \pm 0.013 \) & \(0.472 \pm 0.033 \) & \(0.633 \pm 0.006 \) & \(0.561 \pm 0.005 \)  \\ 
\hline
CVRL~\cite{CVRL}   & 12 (max)    & \(0.598 \pm 0.006 \) & \(0.573 \pm 0.001 \) & \(0.627 \pm 0.004 \) & \(0.526 \pm 0.004 \) & \(0.694 \pm 0.001 \) & \(0.604 \pm 0.002 \)  \\
Ours     & 12          & \(0.590 \pm 0.006 \) & \(0.607 \pm 0.003 \) & \(\mathbf{0.653 \pm 0.009}\)  & \(0.554 \pm 0.011 \) & \(\mathbf{0.709 \pm 0.010 }\) & \(0.622 \pm 0.002 \)  \\ 
Ours     & 128 (max)   & \(\mathbf{0.605 \pm 0.010} \) & \(\mathbf{0.618 \pm 0.000} \) & \(0.630 \pm 0.008 \)  & \(\mathbf{0.619 \pm 0.009} \) & \(0.705 \pm 0.000 \) & \(\mathbf{0.635 \pm 0.003} \)  \\ 
\hline
\end{tabular}
}
\label{tab:64GB}
\end{table*}

Regarding training time, we observe that the training time of CVRL is almost equal to the temporal stage training in our method, where the spatial encoder is frozen. The similarity in training time is largely attributed to the substantial amount of time spent on video loading and preprocessing operations. However, since our method requires an additional spatial stage training period, it results in a longer total training duration than CVRL (approximately twice as long).

\subsection{Analysis}

\begin{table*}[t]
\renewcommand\arraystretch{1.5}
\centering
\caption{Ablation on different pre-training stages (AUCROC)}
\resizebox{\linewidth}{!}{
\setlength{\tabcolsep}{10pt}
\begin{tabular}{ccccccc} 
\hline
Methods   & Fold1        & Fold2   & Fold3    & Fold4     & Fold5    & Mean       \\ 
\hline
w/o pre-training & \(0.554 \pm 0.013 \) & \(0.573 \pm 0.007 \) & \(0.573 \pm 0.013 \) & \(0.472 \pm 0.033 \) & \(0.633 \pm 0.006 \) & \(0.561 \pm 0.005 \)  \\ 
\hline
spatial only & \(\mathbf{0.606 \pm 0.008} \) & \(0.609 \pm 0.004 \) & \(0.621 \pm 0.005 \) & \(0.581 \pm 0.003 \) & \(0.691 \pm 0.010 \) & \(0.622 \pm 0.004 \)  \\ 
temporal only   & \(0.600 \pm 0.006 \) & \(0.573 \pm 0.001 \) & \(0.626 \pm 0.003 \) & \(0.545 \pm 0.036 \) & \(0.679 \pm 0.011 \) & \(0.604 \pm 0.008 \)  \\ 
two-stage (Ours)      & \(0.605 \pm 0.010 \) & \(\mathbf{0.618 \pm 0.000} \) & \(\mathbf{0.630 \pm 0.008} \)  & \(\mathbf{0.619 \pm 0.009} \) & \(\mathbf{0.705 \pm 0.000 }\) & \(\mathbf{0.635 \pm 0.003} \)  \\ 
\hline
\end{tabular}
}
\label{tab:stage}
\end{table*}

\noindent\textbf{Ablation on training stages.}
We validate the effectiveness of each spatial and temporal pre-training stage in Table~\ref{tab:stage}. The results clearly show that pre-training improves performance compared to no pre-training. Both spatial and temporal pre-training individually enhance performance, with spatial pre-training showing a greater improvement than temporal pre-training. This suggests that visual information is more related to qualifying embryos than temporal information. Finally, the highest performance is achieved when both pre-training stages are used together, justifying the efficacy of our two-stage pre-training method.

\begin{table*}[t]
\renewcommand\arraystretch{1.5}
\centering
\caption{Ablation on different augmentations and sampling methods}
\resizebox{\linewidth}{!}{
\setlength{\tabcolsep}{14pt}
\begin{tabular}{cccc|c|c|c} 
\hline
\multicolumn{5}{c|}{Augmentation}                                                                                                           & \multirow{2}{*}{Sampling method} & \multirow{2}{*}{Mean AUCROC}      \\ 
\cline{1-5}
Flip                      & Rotation                  & Brightness                & Contrast                   & Temporally-consistent      &                                  &                                \\ 
\hline
$\surd $ & $\surd $ & $\surd $ & $\surd $  &  & Uniform sampling                    & \(0.597 \pm0.003\)  \\
$\surd $ & $\surd $ & $\surd $ & $\surd $  & $\surd $  & Random sampling                  & \(0.605 \pm0.005 \) \\
$\surd $ & $\surd $ &  &  & $\surd $  & Uniform sampling                    & \(0.631 \pm0.004\)  \\
$\surd $ & $\surd $ & $\surd $ & $\surd $  & $\surd $  & Uniform sampling                    & \(\mathbf{0.635 \pm0.003}\)  \\
\hline
\end{tabular}
}
\label{tab:augmentation}
\end{table*}

\noindent\textbf{Ablation on augmentations.} 
We conduct an ablation study to evaluate the effectiveness of different augmentation techniques in our pre-training method, as summarized in Table~\ref{tab:augmentation}. For spatial augmentations, we implement four augmentations: random rotation, flip, brightness, and contrast jitter. We intentionally avoid augmentations that could significantly alter the appearance of embryos. Additionally, we consider the temporal consistency of the augmentations, deciding whether to apply the same type of augmentation consistently across frames within a single video or to vary the augmentations randomly for each frame. Finally, we compare two frame sampling methods: random sampling and uniform sampling.

The results clearly show the importance of temporally-consistent augmentation in our method. By comparing the first row with the subsequent rows in Table~\ref{tab:augmentation}, it is evident that applying per-frame augmentation without maintaining consistency among frames within a video complicates temporal alignment. Unlike image matching, our loss function is designed to align videos, making temporally-consistent augmentation more appropriate for our approach.

We compare different sampling methods by examining the second row against the remaining rows. The results show that uniform sampling is more effective than random sampling. In embryo videos, the timing of development can be an important indicator of embryo viability. Randomly sampling frames can introduce noise, disrupting the accurate timing of embryo development. Therefore, we chose uniform sampling as our frame sampling method.

Lastly, we compare different spatial augmentations and find that rotation and flipping are important for spatial-stage pre-training.  Embryo videos retain their identity when flipped or rotated, making them suitable for these geometric transformations. Additionally, we applied brightness and contrast augmentations to introduce stronger augmentations. The results indicate that these additional augmentations contribute to improved performance.

\begin{figure*}[t]
\centering
\includegraphics[width=\linewidth,height=0.411\linewidth]{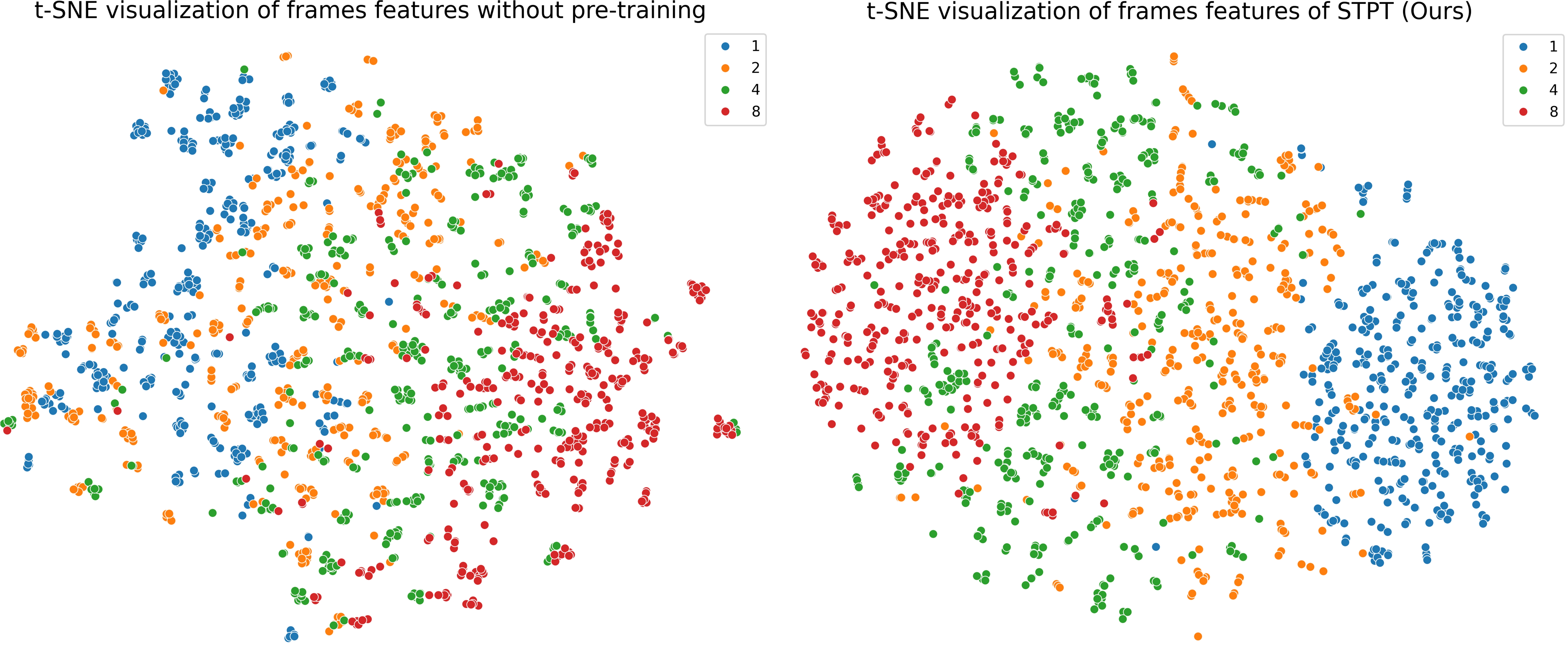}
\caption{T-SNE visualization of frame features before and after pre-training. Each point represents a frame, with different colors indicating the number of cells in each frame. We select frames containing 1, 2, 4, and 8 cells. For each developmental stage, 100 frames are randomly selected for visualization.}
\label{tsne}
\end{figure*}

\noindent\textbf{Feature distribution visualization.} 
To better visualize how pre-training helps the model learn spatial semantic information, we use T-SNE~\cite{tsne} to present the distribution of frame embeddings before and after the spatial stage pre-training. For each developmental stage, 100 frames are randomly selected for visualization. In Fig.~\ref{tsne}, each point represents a frame, with different colors indicating the number of cells in each frame. We select frames containing 1, 2, 4, and 8 cells for distribution visualization, as other stages are rare due to fast transitions. 

The T-SNE plots show a clear difference in distributions between before and after the pre-training stage. It's important to note that the pre-training is self-supervised, meaning the model learns solely from videos without any annotated semantic information. After pre-training, the overall distribution shows a clearer distinction between classes, implying that the model gains an understanding of developmental stages during the pre-training phase.

\section{Conclusion}

In this study, we introduce a Spatial-Temporal Pre-Training (STPT) method to enhance the performance of embryo viability prediction from time-lapse videos. Our method effectively addresses the challenges of memory-efficient training and temporal misalignment in embryo videos by separating the spatial and temporal training stages. 
Our analysis shows that incorporating both spatial and temporal pre-training significantly improves the model's ability for embryo viability prediction. The proposed method is also robust to outliers, attributed to the self-augment and alignment technique, which is important in the medical domain where data variability is common. The proposed STPT method will serve as an effective pre-training approach for embryo viability prediction, particularly when the collected data is limited and contains high variability and outliers.

\section{Acknowledgements}

This work is supported by National Institutes of Health grant R01HD104969.

\end{document}